%% file: nips_2018.tex
\documentclass{article}

\usepackage[preprint]{nips_2018}

\usepackage[utf8]{inputenc} 
\usepackage[T1]{fontenc}    
\usepackage{hyperref}       
\usepackage{url}            
\usepackage{booktabs}       
\usepackage{amsfonts}       
\usepackage{nicefrac}       
\usepackage{microtype}      
\usepackage{amsmath}
\usepackage{amsfonts}
\usepackage{tikz}
\usetikzlibrary{fit,positioning}
\usepackage{relsize}
\usepackage{algpseudocode}
\usepackage{algorithm}
\usepackage{float}
\usepackage{caption}
\usepackage{subcaption}
\usepackage[export]{adjustbox}
\usepackage{setspace}
\def\SP#1{\textsuperscript{{#1}}}

\title{Unveiling the semantic structure of text documents using paragraph-aware Topic Models}

\author{
  Sim\'on Roca-Sotelo  \\
  Dpt Signal Theory and Communications\\
  Universidad Carlos III de Madrid\\
 Spain \\
  \texttt{sroca@ing.uc3m.es} \\
  \And Jer\'onimo Arenas-Garc\'ia \\
  Dpt. Signal Theory and Communications\\
  Universidad Carlos III de Madrid\\
 Spain \\
 \texttt{jeronimo.arenas@uc3m.es}
}

\begin{document}

\maketitle

\begin{abstract}
Classic Topic Models are built under the Bag Of Words assumption, in which word position is ignored for simplicity. Besides, symmetric priors are typically used in most applications. In order to easily learn topics with different properties among the same corpus, we propose a new line of work in which the paragraph structure is exploited. Our proposal is based on the following assumption: in many text document corpora there are formal constraints shared across all the collection, e.g. sections. When this assumption is satisfied, some paragraphs may be related to general concepts shared by all documents in the corpus, while others would contain the genuine description of documents. Assuming each paragraph can be semantically more general, specific, or hybrid, we look for ways to measure this, transferring this distinction to topics and being able to learn what we call specific and general topics. Experiments show that this is a proper methodology to highlight certain paragraphs in structured documents at the same time we learn interesting and more diverse topics.   
\end{abstract}

  \section{Introduction} 
\input{Introduction.tex}

\section{Background}
\input{background.tex}

\section{Paragraph-aware LDA}
\input{parlda.tex}

\section{Experiments}

\input{experiments.tex}

\section{Conclusions and Further work}

In this paper we have identified a specific set of corpora in which changing the semantic unit from words to paragraphs becomes helpful. We have shown with a simple model the benefits of this approach when the proposed generative model is met -- in a synthetic dataset--, and also we are satisfied with the resultant topics and structures learned from real datasets. Highlighting paragraphs seems a reasonable way to tell Topic Modeling algorithms where they should put their efforts into learning high quality topics.

Future analysis should lead to improving the way in which a paragraph is classified as relevant. This will require the use of hand-labeled datasets and new metrics, and also more adapted inference models. Once this paragraph characterization is well studied, users of this approach should be able to give the model more information about the type of topics they are looking for. 

\subsubsection*{Acknowledgments}

This work has been partially supported by MINECO Projects TEC2014-52289-R and TEC2017-83838-R. Sim\'on Roca-Sotelo has received financial support through the “la Caixa” Fellowship Grant for Doctoral Studies, “la Caixa” Banking Foundation, Barcelona, Spain.
\bibliographystyle{humannat}

 \bibliography{mybib}

\end{document}

%% file: Introduction.tex
Topic Modeling refers to a popular set of algorithms that has been widely used for inferring topics or themes --defined by probability vectors over words-- present on collections of text documents of any kind --news, books, scientific articles, patents, e-mails, biological data or even tweets--. Topic modeling quickly became popular after the first models were proposed \citep{deerwester1990indexing}, \citep{blei2003latent}. Since then, a huge amount of related contributions appeared, on the one hand applying these models to specific problems and challenges, and on the other hand exploring more complex models adding capabilities to their ancestors. These new models try to overcome assumptions from the old ones. For instance, Latent Dirichlet Allocation (LDA) \citep{blei2003latent} assumed word-independence (Bag of Words), a pre-fixed number of topics, etc., simplifications that facilitate the inference, both in complexity and computation costs, achieving a fairly good performance on interpretability and many other tasks. However, this model ignores the fact that the ordering of the words contains valuable information. The interested reader could find interesting reviews of previous and current algorithms in \citep{blei2012probabilistic} and \citep{INR-030}.

In this paper, the following assumption is analysed: in some corpora, paragraphs are the basic unit of semantic information. More specifically, when the assumption holds, documents are structured in paragraphs, with some paragraphs being semantically more meaningful than others. To keep it simple, we distinguish two types of paragraphs: {\em general}, that are semantically similar to other general paragraphs contained in most documents of the corpora, and {\em specific}, that contain the most important (and discriminative) semantic information of the documents. Consequently, we distinguish also between \textit{general topics}, those corpus-related which appear in the majority of documents, and \textit{specific ones},  which contain the information which genuinely describes a subset of documents. We claim that, when the assumption is met, our model is able to learn document structure and better quality topics.

The rest of the paper is organized as follows: in Section 2, we will briefly summarize previous contributions based on similar approaches or objectives, pondering how they differ from ours. In Section 3, we introduce a simple generative model for documents which allow us to derive a Gibbs Sampling based inference matching our purpose. In Section 4, we apply this inference on synthetic and real datasets, proving that this model works better when the assumptions are met, and exploring topics learned in real datasets. Finally, in Section 5 we discuss the findings of the experiments in the previous Section and suggest future applications and lines of work.

%% file: background.tex
In this section, we briefly introduce some existing techniques that have already produced algorithms related to this work. This review establishes the grounds for better understanding the motivation of our work. 

\subsection{Previous work}

There are previous models in literature proposing learning topics separately depending on how document-specific they are. In \cite{chemudugunta2007modeling} they propose a model that considers words coming from three possible topical sources --a corpus shared background distribution, a document-specific distribution, or a corpus-specific set of topics--. The model tries to isolate stopwords and other non-relevant words in the background distribution, while the rest of words are modeled depending on how often they are shared across documents. This improved the ability of matching queries, specially for low frequency words, since they matched those of document-specific topics. The assumed generation model presents three paths for generating words, and it is controlled by an additional latent variable, a Multinomial distribution acting as a switch. However, their choice is done word per word and is therefore subject to the limitations of the Bag of Words assumption. In contrast, our work is based on the assumption that there exists certain structure in the document at the paragraph level, so that general or specific words tend to occur (at least in certain corpora) separately on different paragraphs. Then, algorithms aware of this structure will produce better topics, and as a subproduct find the most semantically relevant paragraphs of each document. 

 \cite{haghighi2009exploring} considers also a background distribution, and content topics that may be general (for a collection) or specific (for a document). A sentence may contain background words and specific words, but all the specific words of the sentence must belong to the same topic. Topic transition between sentences is modeled with a Hidden Markov Model (HMM), with a high probability of keeping the same topic. Unlike this work, our model allows several topics both in specific paragraphs and background ones.

Finally, other authors have suggested models in which text segments share common attributes, mainly sentences. For instance, in \citet{gruber2007hidden} each sentence is assumed to share the same topic for all its words. Transitions between topics are also modeled as a HMM. This permits identifying sections in documents like scientific articles. In \citet{balikas2016topic} authors suggest a model in which topics are learned at a sentence level. They revisit a typical Gibbs Sampling inference to consider how to compute the probability of a full sentence belonging to a topic. This model differs from ours because it assumes one topic per sentence. In ours each word can be sampled from different topics instead.

\subsection{Motivation of this contribution}

We assume that there are certain text document corpora in which, due to several factors like formal structure --job offers, grants proposals, patents, articles, etc-- it is worth to model paragraphs separately. We expect these documents to manifest, at paragraph level, contextual information related to the corpus itself on one hand, and more specific and distinctive information (i.e., document-specific) on the other hand. In the first ones, we learn topics describing the corpus structure and general content, while in the second ones we discover more specific topics. We provide a model that allows detecting these paragraphs when the model is met. In the end, our motivation is learning higher quality topics in general, disentangling them when possible. In addition, the model unveils the structure of the document highlighting the most specific paragraphs. It is worth mentioning that choosing paragraphs as the semantic unit level is a trade-off between sentences and longer segments (e.g., sections). Obviously, the model would remain unchanged if the text span were changed.

%% file: parlda.tex
\subsection{Mathematical Notation and Generative model}

Table \ref{tab1} summarizes the most important variables and parameters that are necessary for the presentation of our model. 

{\renewcommand{\arraystretch}{1.3}
\begin{table}[t]
  \centering
  \begin{tabular}{ll}
    \toprule
Variable & {Description} \\    
\midrule
$\vec{w}$ & Observed words in the text collection. \\
$\vec{z}$ & Topic assignments for each of the words. \\
$\vec{\theta}_{1}^{(d)} (\vec{\theta}_{0}^{(d)})$ & Vector with the specific (general) topic distribution of document $d$. \\
$\alpha_{1} (\alpha_{0})$ & Hyperparameter for prior $\vec{\theta}_{1}^{(d)} (\vec{\theta}_{0}^{(d)})$ \\
$\vec{x}$ & Paragraph assignments: $x_{i}=1$ if the paragraph is specific, and $x_{i}=0$ if general. \\
$\vec{\psi}^{(d)}$ & Proportion of specific and general paragraphs in a document $d$. \\ 
$\gamma$ & Hyperparameter for prior $\vec{\psi}^{(d)}$  \\
$\vec{\phi}_{1,k} (\vec{\phi}_{0,k})$ & Vector containing the vocabulary distribution for specific (general) topic $k$. \\
 $K_{1} (K_{0})$ & Total number of specific (general) topics. In general, $K_{1} \neq K_{0}$  \\
$\beta_{1} (\beta_{0})$ & Hyperparameter for prior $\vec{\phi}_{1} (\vec{\phi}_{0})$ \\
$m$ & Proportion of words generated by specific topics in a specific paragraph \\
    \bottomrule
  \end{tabular}
    \caption{Variables and parameters for the generative model.\label{tab1}}
  \label{params}
\end{table}

To be more specific, the generative model corresponding to our model can be summarized as follows:

\begin{enumerate}
\item Specific (general) topics are sampled as: \\ 
{\setstretch{2}
$\vec{\phi}_{1,k} \sim Dir_{V}(\beta_{1}), k=1,...,K_{1}$ \\
 $\vec{\phi}_{0,k} \sim Dir_{V}(\beta_{0}), k=1,...,K_{0}$ }
\item For each document,
\begin{enumerate}
\item Topic proportions for specific and general topics are sampled: \\
{\setstretch{2}
$\vec{\theta}_{1}^{(d)} \sim Dir_{K_{1}}(\alpha_{1}<<1)$ (narrower in documents) \\
$\vec{\theta}_{0}^{(d)}  \sim Dir_{K_{0}}(\alpha_{0} \geq 1)$ (wider in documents) }
\item The proportion of specific and general topics is obtained: \\
{\setstretch{2}
$\vec{\psi}^{(d)} \sim Dir_2(\gamma)$ }
\item For each paragraph in the document, 
\begin{itemize}
\item Choose whether the paragraph is specific ($x_{i}=1$) or general ($x_{i}=0$): \\
{\setstretch{2}
$x_{i} \sim Ber(\psi_{x})$ }
\item For each word in the paragraph: 
\begin{itemize}
\item if $x_{i}=0$, sample the general topic and word from the selected topic: \\
$z_{i} \sim Mult(\vec{\theta}_{0}^{(d)})$ and $w_{i} \sim Mult(\vec{\phi}_{0,z_{i}})$.
\item if $x_{i}=1$:
\begin{itemize}
\item Sample if the word comes from an specific or general topic ($x^{(w_{i})}=1$ or $x^{(w_{i})}=0$), using $m$.
\item Sample the topic from $\vec{\theta}_{1}^{(d)}$ or $\vec{\theta}_{0}^{(d)}$, and sample the word from the selected topic: $z_{i} \sim Mult(\vec{\theta}_{x_{w}}^{(d)})$ and $w \sim Mult(\vec{\phi}_{x_{w},z_{i}})$.
\end{itemize}

\end{itemize}
\end{itemize}
\end{enumerate}
\end{enumerate}

The additional variables that are introduced with respect to standard LDA are due to the following differences of our model --see also the graphical model in Fig. \ref{new}:

\begin{itemize}
\item An extra plate is included to reflect the paragraph level. For each paragraph we have included an additional binary variable $x$ that can take values $0$ and $1$ for general and specific paragraphs, respectively. This way, specific topics and general topics are sampled separately, as well as the topic-document proportions of each kind.
\item Even if a paragraph is described as specific, assuming all its words are going to be specific is unrealistic. For that, we introduce a mixing probability $m$ allowing an arbitrary small proportion of general words. Inferring $m$ is out of the scope of this work, and its value is fixed before the training stage.
\item All the variables whose output is a probability distribution have a Dirichlet prior, whereas all the assignments are modeled as a Multinomial distribution (this choice ensures conjugacy). 
\item All priors are symmetric, i.e., hyperparameters $\alpha_{1}$, $\alpha_{0}$, $\beta_{1}$, $\beta_{0}$ and $\gamma$ are considered scalar values.
\item $\alpha_{1} << 1$, forces to learn specific topics which, for each document, only a few will appear.
\item $\alpha_{0} \geq 1$ favours all general topics to appear in all documents.
\end{itemize}

\begin{figure}[t]
\centering
\begin{tikzpicture}
\tikzstyle{main}=[circle, minimum size = 8mm, thick, draw =black!80, node distance = 13mm,scale=0.8]
\tikzstyle{connect}=[-latex, thick]
\tikzstyle{box}=[rectangle, draw=black!100]
  \node[main, fill = white!100] (alpha1) [label=left:$\alpha_{1}<<1$] { };
   \node[main] (theta1) [right=of alpha1,label=below:$\theta_{1}$] { };
    \node[main] (z) [right=of theta1,label=below:z] {};
      \node[main, fill = black!10] (w) [right=of z,label=below:w] { };
    \node[main, fill = white!100] (alpha0) [above=4.5mm of alpha1, label=left:$\alpha_{0}\geq1$] { };
  \node[main] (theta0) [right=of alpha0,label=below:$\theta_{0}$] { };
   \node[main] (x) [above=6.5mm of z,label=right:$x$] { };
      \node[main] (psi) [above=4.5mm of theta0,label=below:$\psi$] { };
      \node[main] (gamma) [above=4.5mm of alpha0,label=left:$\gamma$] { };
    \node[main] (phi1) [right=of w,label=below:$\phi_{1}$] { };
  \node[main] (phi0) [above=of phi1,label=below:$\phi_{0}$] { };
    \node[main] (beta1) [right=of phi1,label=right:$\beta_{1}$] { };
  \node[main] (beta0) [above=of beta1,label=right:$\beta_{0}$] { }; 

  \path (alpha0) edge [connect] (theta0)
(alpha1) edge [connect] (theta1)        
        (theta0) edge [connect] (z)
        (theta1) edge [connect] (z)
		(z) edge [connect] (w)
		(phi0) edge [connect] (w)
		(beta0) edge [connect] (phi0)
		(beta1) edge [connect] (phi1)
		(gamma) edge [connect] (psi)
(psi) edge [connect] (x)		
		(x) edge [connect] (z)
		(phi1) edge [connect] (w);
  \node[rectangle, inner sep=0mm, fit= (z) (w),label=below right:N, xshift=9.5mm] {};
  \node[rectangle, inner sep=4.4mm,draw=black!100, fit= (z) (w)] {};
  \node[rectangle, inner sep=4.6mm, fit= (z) (w),label=below right:M, xshift=9mm] {};
  \node[rectangle, inner sep=9mm, draw=black!100, fit = (theta0) (theta1) (z) (w) (psi)] {};
   
    \node[rectangle, inner sep=5.5mm, draw=black!100, fit = (x) (z) (w)] {};
        \node[rectangle, inner sep=5.8mm, fit = (x) (z) (w), label=above right: P,xshift=-5.5mm,yshift=-5mm] {};

  \node[rectangle, inner sep=5mm, draw=black!100, fit = (phi0),xshift=3mm] {};
\node[rectangle, inner sep=5.2mm, fit = (phi0),label=below right:$K_{0}$, xshift=-4mm,yshift=5mm] {};  
  \node[rectangle, inner sep=5mm, draw=black!100, fit = (phi1),xshift=3mm] {};
  \node[rectangle, inner sep=5.2mm, fit = (phi1),label=below right:$K_{1}$, xshift=-4mm,yshift=5mm] {};  

\end{tikzpicture}
\caption{Paragraph LDA graphical model}
\label{new}
\end{figure}
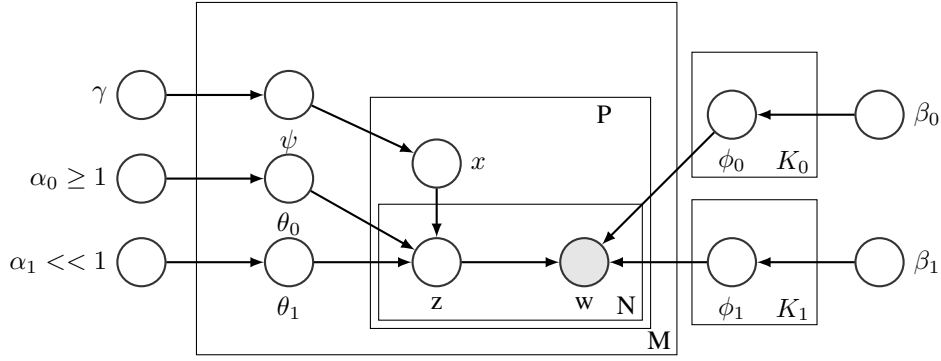

%
\subsection{Inference}

Inference is based on Collapsed Gibbs Sampling (see \cite{griffiths2004finding}, \cite{textanalysis}). As for LDA, the factorized joint probability of the model is used to obtain full conditionals of the hidden variables we want to estimate. In LDA, $\vec{\theta}^{(d)}$ and $\vec{\phi}_{k}$ are integrated out (collapsed) since they can be estimated as statistics of topic assignments $\vec{z}$. In our model, $\vec{\psi}^{(d)}$ is collapsed too, and the factorized joint probability is the following:

\begin{equation}
p(\vec{z},\vec{x}, \vec{w}) = p(\vec{w} | \vec{z}, \vec{\beta}) p(\vec{z} | \vec{x}, \vec{\alpha}) p(\vec{x} | \gamma)
\end{equation}

The full conditional for $\vec{x}$ is obtained following a similar approach. First, we integrate out $\vec{\psi}^{(d)}$, resulting in the following Dirichlet-Multinomial distribution over $\vec{x}$:

\begin{equation}
p(\vec{x} | \gamma) = \mathlarger{\frac{\Delta(\vec{n}_{x}'+\gamma)}{\Delta(\gamma)}} 
\end{equation}
where $\vec{n}_{x}'$ expresses the occurrences of specific and general paragraphs in the corpus.

Up to this point, the procedure for the two other factors in the joint probability is the same. However, when obtaining the full conditionals, the Dirichlet-Multinomials over $\vec{w}$ and $\vec{z}$ became proportional to Gamma function quotients. In LDA, we can implicitly make use of the fact that $\Gamma(x+1) = x\Gamma(x)$ and $\vec{z} = \{z_{i},\vec{z}_{\neg i}\}$, and the full conditional for $z_{i}$ is approximated by:

\begin{equation}
p(z_{i}=k|\vec{z_{\neg i}}, \vec{w}) \propto (n_{m,\neg i}^{(k)} + \alpha_{k}) \frac{n_{k,\neg i}^{(t)} + \beta_{k}}{\sum^{V}_{t=1} n_{k,\neg i}^{(t)} + \beta_{k}} 
\end{equation} 

where:
\begin{itemize}
\item $n_{m,\neg i}^{(k)}$ counts how many times the topic $k$ appears in document $m$, ignoring the current assignment $i$; 
\item $n_{k,\neg i}^{(t)}$ refers to how many times the term $t$ was sampled from topic $k$ in other assignments.
\end{itemize}
For each paragraph assignment $x$, however, we are now counting the number of words in documents belonging to each type of paragraphs. When obtaining the full conditional, we have to take into account that changing one paragraph assignment will affect more than one word --$\vec{w}_{s} = \{\vec{w}_{s,p},\vec{w}_{s,\neg p}\}$, where $\vec{w}_{s,p}$ express the words assigned to a certain type of paragraph-- and the recursion rule for Gamma function is applied in general more than once. This leads to the following full conditional, which is analogous to sentence-topics in \cite{balikas2016topic}. 

\begin{equation}
p(x_{p} = s | \vec{w}, \vec{x}_{\neg p}, \gamma) \propto \frac{ \prod_{t \in p} (n_{s, \neg p}^{(t)} + \gamma)...(n_{s, \neg p}^{(t)} + \gamma +( n_{s, p}^{(t)}-1 )) }{(\sum_{w\in V}(n_{s, \neg p}^{(w)} + \gamma))...(\sum_{w\in V}n_{s, \neg p}^{(w)} + \gamma +( n_{s, p}^{(w)}-1 )) }
\end{equation} 
where $n_{s, p}^{(t)}$ expresses how many times term $t$ appears in paragraph $p$, assuming this one being $s\in\{0,1\}$.

%
%
%
%
%
%
%
%
%
%

%% file: experiments.tex
In this section, we present different results on synthetic and real datasets in order to validate our proposal. We consider two sets of experiments:

\begin{itemize}
\item Firstly, we construct a synthetic corpus using the generative model from the previous section. Since we create the corpus, we know the real labels for the specific and general paragraphs. These first experiments seek to validate the inference scheme and to show that, when our assumption about document structure holds, we can gain over usage of standard LDA both w.r.t. the quality of learned topics and the ability to discriminate between both kinds of paragraphs even when LDA is used together with a classifier that is given the true labels.
\item Secondly, we explore a real dataset --a subset of USPTO patents\footnote{collected from \url{https://bulkdata.uspto.gov/data/patent/grant/redbook/fulltext/2017/}} --. The intention here is to study if the proposed model can obtain better topics and identify relevant paragraphs in a real dataset.
\end{itemize}

\subsection{Experiments on synthetic dataset}

\begin{table}[t]
  \centering
  \begin{tabular}{ccccccc}
    \toprule
    Docs (test) & Paragraphs (test) & Words(test) & $K_{0}(K_{1})$ & $V$ & $\alpha_{0}(\alpha_{1})$ & $\beta_{0}(\beta_{1})$  \\
    \midrule
    3000(500) & 62627 (12439)  & 2191252 (434052) & 10 (30) & 5000 & 2 (0.1) & 0.1(0.1)    \\
    \bottomrule
  \end{tabular}
    \caption{Attributes for Synthetic dataset generation.}
  \label{syn-table}
\end{table}

Table \ref{syn-table} contains the parameters that were used for generating this corpus, namely: the vocabulary size $V$, the number of topics $K_{1}$ and $K_{2}$, prior hyperparameters and number of documents, paragraphs and words generated, both for a train set and a test set . In addition, to make the problem more difficult some noise was added varying the $m$ parameter, that is, the proportion of general words in specific paragraphs. Concretely, for each document, $m \sim Uniform(0.2,0.8)$. This may lead to confusing situations in which paragraphs labeled as specific contain only $20\%$ of specific words. Hyperparameter $m$ was set to $0.5$ during the inference. 

In order to prove the capability of our method to identify documents structure, once the corpus was generated, three models were compared w.r.t. their ability to discriminate general and specific paragraphs:
\begin{enumerate}
\item parLDA (our model): after learning topics on the train set, paragraph probabilities were estimated on the test set.
\item LDA+SVM: topics were learned on the train set. Then, topic assignments were sampled on the test set. Characterizing each paragraph with its assignments, half the test set was used with its real labels to train an RBF-kernel SVM, whose parameters $C$ and $\gamma$ were crossvalidated. Probabilities estimates were provided for the final test set.
\item BoW+SVM: instead of topic assignments, BoW vectors from the paragraphs in the training set were used to train a linear SVM, and to obtain probability estimates for the test set. The reason for choosing a linear SVM is that we wanted a ground truth on how separable were the paragraphs before applying Topic Modeling. 
\end{enumerate}

\begin{figure}[t]
\centering
 \includegraphics[scale=0.6]{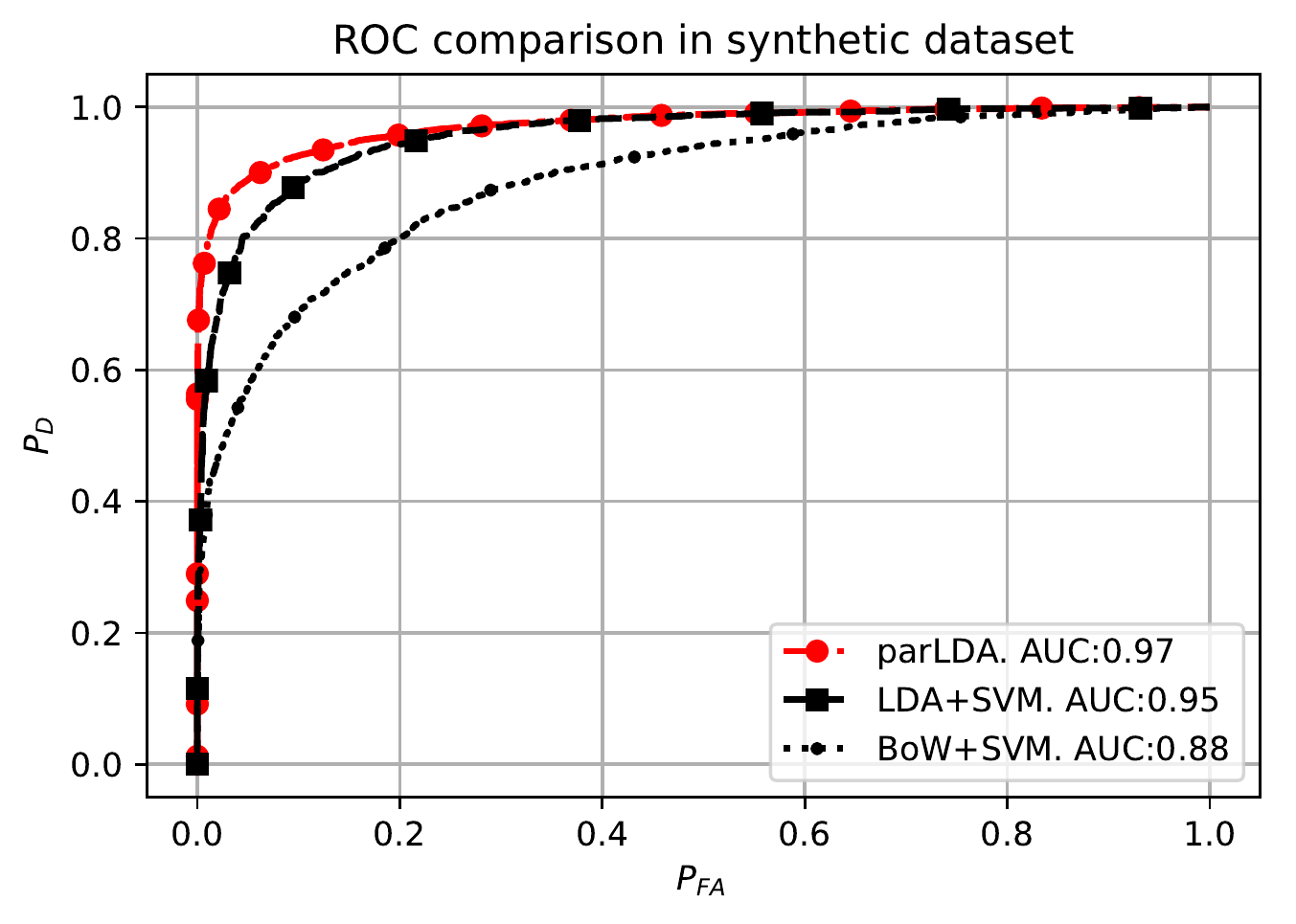}
  \caption{ROC curves for paragraphs detection.}
  \label{fig:ROC}
  \end{figure}

ROC curves for these methods are shown in Fig. \ref{fig:ROC}. It can be observed that parLDA outperforms the other methods, even though true labels are used by LDA+SVM and BoW+SVM while our method is fully unsupervised.

A second experiment was carried out to evaluate the learned topics from parLDA and LDA. Since we had the real topics, we looked for metrics measuring distances/similarities among probability distributions to compare them against the learned topics. We used a Histogram Intersection similarity metric (see \cite{cha2002measuring}) to see how many original topics the algorithms are able to identify and how close the learned topics are from them, comparing our method and LDA. parLDA similarities were slightly superior to those of LDA. This make us think that parLDA typically learns better topics than LDA in this specific scenario where our generative assumption was right. To prove this, we defined a `correctly guessed' topic as the one which coincides in at least 5 words with the one is predicting (in a top-10 representation), to see if these higher similarities lead to better learned topics: parLDA correctly guessed 37 out of 40 topics, whereas LDA guessed 28 out of 40.

\subsection{Experiments on USPTO patents}

In this second scenario we apply our method to a real corpus. Concretely, we selected 3000 patents from the week of January 31\SP{th}, 2017. Table \ref{real-table} shows more information about the corpus and the selected parameters for inference.

\begin{table}[t]
  \centering
  \begin{tabular}{ccccccc}
    \toprule
    Docs & Paragraphs & Words & $K_{0}(K_{1})$ & $V$ & $\alpha_{0}(\alpha_{1})$ & $\beta_{0}(\beta_{1})$  \\
    \midrule
    3000 & 410652  & 8652304 & - (15) & 8323 & 2 (0.1) & $500/V$($200/V$)    \\
    \bottomrule
  \end{tabular}
    \caption{Attributes for USPTO patents experiments.}
  \label{real-table}
\end{table}

Table \ref{real-table} contains the description for this corpus. Now, $K_{0}$, the number of general topics, will be variable. $K_{1}$ is fixed to 15 as we will see below. In order to check which topics our model learns and which paragraphs it highlights, we defined the following experiment:

\begin{enumerate}
\item First, we run a classic LDA algorithm (using the same parameters as those for specific topics in Table \ref{real-table}) to determine a proper number of topics $K^{*}$ to start comparisons. We will use for that a Topic Coherence measurement (concretely, one from \cite{roder2015exploring}).
\item Then, when the number of topics $K^{*}$ is fixed, we launch our method to learn $K{1}=K^{*}$ specific topics and $K_{0}$ as an incremental number of general topics ($1,2,3,4,5,10$). This way, we can check if adding new topics in the general topics set helps improving specific topics quality, compared to LDA topics. 
\end{enumerate}

We decided to use here Topic Coherence to measure the topic quality since perplexity has been proved that is not correlated in all cases with human perception (see \cite{chang2009reading}). Some of them are based on co-ocurrences of words in a reference corpus, proving a high correlation with human judgement \citep{lau2014machine}, while there have been interesting proposals based on similarities of the word-vectors of topics \citep{fang2016using}. In addition, some measurements based on the corpus itself have been proposed too \citep{mimno2011optimizing}. 

After looking at the results of some of the abovementioned coherences, including one based in pre-trained in Fasttext word vectors \citep{bojanowski2016enriching}, we observed that most of them have unexpected behaviours when learned topics contain very domain-specific words, not being able to capture some semantic relationships. In the end, we selected the $C_{P}$ measurement in \cite{roder2015exploring}, after providing the most sensible results. Human-correlated automatic coherence measurements are still a challenge to be analyzed in the future. 

\begin{figure}[t]
\begin{subfigure}{.5\textwidth}
  \includegraphics[width=.9\linewidth,center]{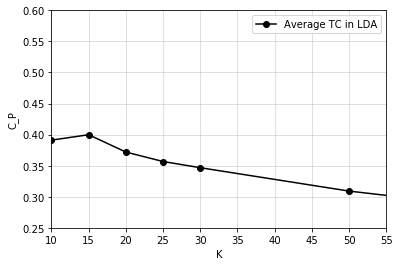}
  \label{fig:sub1}
\end{subfigure}%
\begin{subfigure}{.5\textwidth}
  \includegraphics[width=.9\linewidth,center]{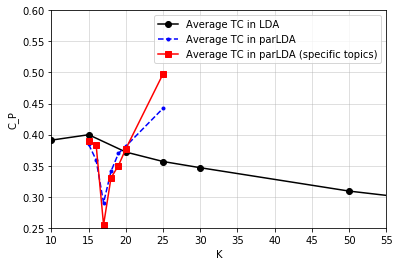}
  \label{fig:sub2}
\end{subfigure}
\caption{Topic Coherence on parLDA topics vs LDA.}
\label{fig:coherences}
\end{figure}

Fig. \ref{fig:coherences} (left) shows that, for $K^{*}=15$, LDA learns a set of topics which, in average, have a higher Topic Coherence. Fig. \ref{fig:coherences} (right) compares that curve with the ones resulting from Topic Coherences for parLDA topics (average of all topics, and average of specific topics only, set to 15). When adding one or two background topics, we can see that coherence is worse. This may be related to how restrictive is to represent general paragraphs with one or two topics. However, the more general topics we add, the higher Coherence is for specific topic, higher than for LDA. This proves that adding general topics in which more general words could fit in, helps in cleaning specific topics, those in which we want to focus. Lastly, Table \ref{topic-table} shows some of the learned topics in this experiment. 

\begin{table}[t]
  \centering
  \resizebox{\columnwidth}{!}{
  \begin{tabular}{c|c|c}
    \toprule
Topic type & $C_{P}$ coherence &Top 10 words \\    \midrule
  Spec. & 0.86 & compound, acid, carbon, solution, atom, solvent, formula, polymer, reaction, alkyl    \\
  Spec. & 0.79 & acid, composition, peptide, agent, enzyme, amino, aqueous, compound, surfactant, ester    \\ \midrule
  Gen. & 0.67 & optical, electrode, voltage, lens, terminal, substrate, transistor, led, magnetic, coil    \\
  Gen. & 0.63 & sensor, vehicle, controller, switch, cell, module, mode, threshold, voltage, battery    \\ \midrule
  LDA & 0.82 & optical, lens, beam, led, wavelength, laser, radiation, emission, mirror, angle    \\
  LDA & 0.71 & temperature, polymer, particle, resin, coating, weight, composition, metal, glass, fiber    \\
    \bottomrule
  \end{tabular}
}
    \caption{Higher coherence topics, comparison between specific, general, and LDA topics.}
  \label{topic-table}
\end{table}

\begin{figure}[t]
\centering
\includegraphics[width=0.8\textwidth]{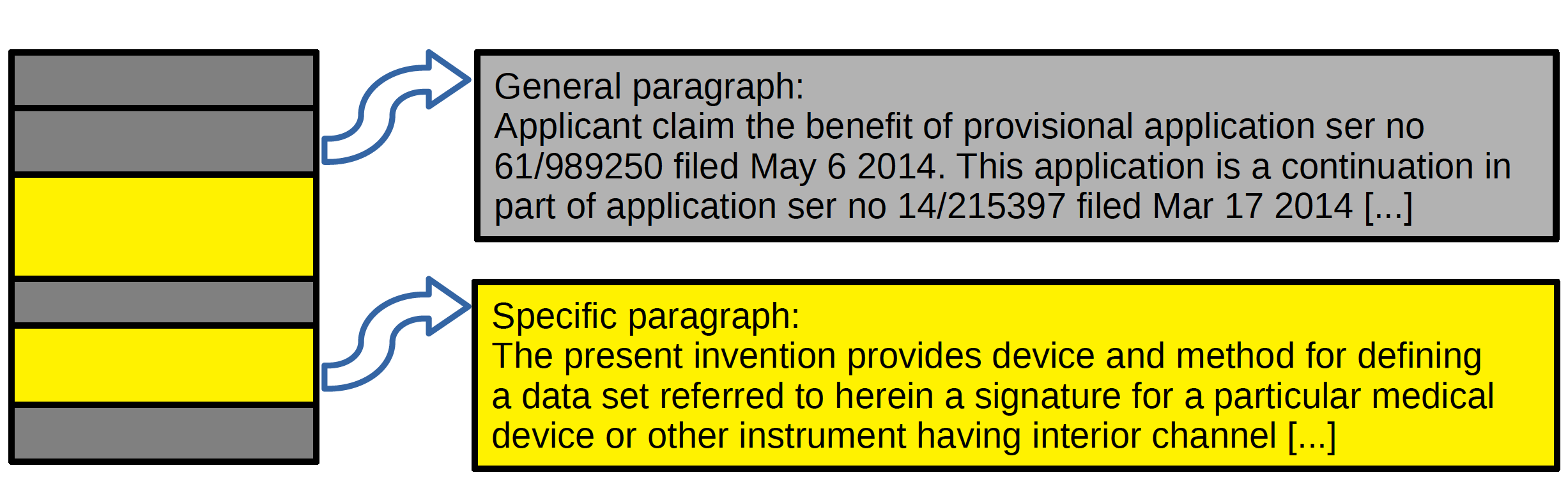}
\caption{Example of highlighted specific and general paragraphs in a patent.}
  \label{paragraphs}
\end{figure}

%% file: nips_2018.bbl
\begin{thebibliography}{}

\bibitem[\protect\astroncite{Balikas et~al.}{2016}]{balikas2016topic}
Balikas, G., M.-R. Amini, and M.~Clausel\leavevmode\nopagebreak\newline 2016.
\newblock On a topic model for sentences.
\newblock In {\em Proceedings of the 39th International ACM SIGIR conference on
  Research and Development in Information Retrieval}, Pp.~ 921--924. ACM.

\bibitem[\protect\astroncite{Blei}{2012}]{blei2012probabilistic}
Blei, D.~M.\leavevmode\nopagebreak\newline 2012.
\newblock Probabilistic topic models.
\newblock {\em Communications of the ACM}, 55(4):77--84.

\bibitem[\protect\astroncite{Blei et~al.}{2003}]{blei2003latent}
Blei, D.~M., A.~Y. Ng, and M.~I. Jordan\leavevmode\nopagebreak\newline 2003.
\newblock Latent dirichlet allocation.
\newblock {\em Journal of machine Learning research}, 3(Jan):993--1022.

\bibitem[\protect\astroncite{Bojanowski et~al.}{2016}]{bojanowski2016enriching}
Bojanowski, P., E.~Grave, A.~Joulin, and
  T.~Mikolov\leavevmode\nopagebreak\newline 2016.
\newblock Enriching word vectors with subword information.
\newblock {\em arXiv preprint arXiv:1607.04606}.

\bibitem[\protect\astroncite{Boyd-Graber et~al.}{2017}]{INR-030}
Boyd-Graber, J., Y.~Hu, and D.~Mimno\leavevmode\nopagebreak\newline 2017.
\newblock Applications of topic models.
\newblock {\em Foundations and Trends® in Information Retrieval},
  11(2-3):143--296.

\bibitem[\protect\astroncite{Cha and Srihari}{2002}]{cha2002measuring}
Cha, S.-H. and S.~N. Srihari\leavevmode\nopagebreak\newline 2002.
\newblock On measuring the distance between histograms.
\newblock {\em Pattern Recognition}, 35(6):1355--1370.

\bibitem[\protect\astroncite{Chang et~al.}{2009}]{chang2009reading}
Chang, J., S.~Gerrish, C.~Wang, J.~L. Boyd-Graber, and D.~M.
  Blei\leavevmode\nopagebreak\newline 2009.
\newblock Reading tea leaves: How humans interpret topic models.
\newblock In {\em Advances in neural information processing systems}, Pp.~
  288--296.

\bibitem[\protect\astroncite{Chemudugunta
  et~al.}{2007}]{chemudugunta2007modeling}
Chemudugunta, C., P.~Smyth, and M.~Steyvers\leavevmode\nopagebreak\newline
  2007.
\newblock Modeling general and specific aspects of documents with a
  probabilistic topic model.
\newblock In {\em Advances in neural information processing systems}, Pp.~
  241--248.

\bibitem[\protect\astroncite{Deerwester et~al.}{1990}]{deerwester1990indexing}
Deerwester, S., S.~T. Dumais, G.~W. Furnas, T.~K. Landauer, and
  R.~Harshman\leavevmode\nopagebreak\newline 1990.
\newblock Indexing by latent semantic analysis.
\newblock {\em Journal of the American society for information science},
  41(6):391.

\bibitem[\protect\astroncite{Fang et~al.}{2016}]{fang2016using}
Fang, A., C.~Macdonald, I.~Ounis, and P.~Habel\leavevmode\nopagebreak\newline
  2016.
\newblock Using word embedding to evaluate the coherence of topics from twitter
  data.
\newblock In {\em Proceedings of the 39th International ACM SIGIR conference on
  Research and Development in Information Retrieval}, Pp.~ 1057--1060. ACM.

\bibitem[\protect\astroncite{Griffiths and
  Steyvers}{2004}]{griffiths2004finding}
Griffiths, T.~L. and M.~Steyvers\leavevmode\nopagebreak\newline 2004.
\newblock Finding scientific topics.
\newblock {\em Proceedings of the National academy of Sciences}, 101(suppl
  1):5228--5235.

\bibitem[\protect\astroncite{Gruber et~al.}{2007}]{gruber2007hidden}
Gruber, A., Y.~Weiss, and M.~Rosen-Zvi\leavevmode\nopagebreak\newline 2007.
\newblock Hidden topic markov models.
\newblock In {\em Artificial intelligence and statistics}, Pp.~ 163--170.

\bibitem[\protect\astroncite{Haghighi and
  Vanderwende}{2009}]{haghighi2009exploring}
Haghighi, A. and L.~Vanderwende\leavevmode\nopagebreak\newline 2009.
\newblock Exploring content models for multi-document summarization.
\newblock In {\em Proceedings of Human Language Technologies: The 2009 Annual
  Conference of the North American Chapter of the Association for Computational
  Linguistics}, Pp.~ 362--370. Association for Computational Linguistics.

\bibitem[\protect\astroncite{Heinrich}{2005}]{textanalysis}
Heinrich, G.\leavevmode\nopagebreak\newline 2005.
\newblock Parameter estimation for text analysis.

\bibitem[\protect\astroncite{Lau et~al.}{2014}]{lau2014machine}
Lau, J.~H., D.~Newman, and T.~Baldwin\leavevmode\nopagebreak\newline 2014.
\newblock Machine reading tea leaves: Automatically evaluating topic coherence
  and topic model quality.
\newblock In {\em Proceedings of the 14th Conference of the European Chapter of
  the Association for Computational Linguistics}, Pp.~ 530--539.

\bibitem[\protect\astroncite{Mimno et~al.}{2011}]{mimno2011optimizing}
Mimno, D., H.~M. Wallach, E.~Talley, M.~Leenders, and
  A.~McCallum\leavevmode\nopagebreak\newline 2011.
\newblock Optimizing semantic coherence in topic models.
\newblock In {\em Proceedings of the conference on empirical methods in natural
  language processing}, Pp.~ 262--272. Association for Computational
  Linguistics.

\bibitem[\protect\astroncite{R{\"o}der et~al.}{2015}]{roder2015exploring}
R{\"o}der, M., A.~Both, and A.~Hinneburg\leavevmode\nopagebreak\newline 2015.
\newblock Exploring the space of topic coherence measures.
\newblock In {\em Proceedings of the eighth ACM international conference on Web
  search and data mining}, Pp.~ 399--408. ACM.

\end{thebibliography}
